\documentclass[12pt]{iopart}

\usepackage{graphicx}
\usepackage{float}

\usepackage{natbib}
\setcitestyle{comma}
\bibpunct{(}{)}{ }{a}{}{,}
\usepackage{booktabs}
\usepackage{threeparttable}
\usepackage{siunitx}
\usepackage{multirow}
\usepackage{subcaption}

\begin{document}

% --- TITRE ---
\title[ML-Based ICS Recovery for UHR-PET]{Machine Learning-Based Inter-Crystal Scatter Recovery for Ultra-High Resolution PET Imaging}

\author{Alexandre Bernier$^{1,2,3}$, Roger Lecomte$^{2,3,5,6}$, Jean-Baptiste Michaud$^{1,2,3}$}
\address{$^1$Department of Electrical \& Computer Engineering, Université de Sherbrooke, Sherbrooke, QC, Canada}
\address{$^2$Interdisciplinary Institute for Technological Innovation, Sherbrooke, QC, Canada}
\address{$^3$Sherbrooke Molecular Imaging Center of CRCHUS, Sherbrooke, QC, Canada}
\address{$^4$Institut Fresnel, CNRS UMR7249, Aix Marseille Univ, Centrale Med, 13013 Marseille, France}
\address{$^5$Department of Medical Imaging and Radiation Sciences, Université de Sherbrooke, Sherbrooke, QC, Canada}
\address{$^6$Imaging Research \& Technology  (IR\&T) Inc., Magog, QC, Canada}
\ead{alexandre.bernier2@usherbrooke.ca}

\maketitle

% --- CONTENU ---
\begin{abstract}
Inter-crystal scatter (ICS) events pose a significant challenge in ultra-high-resolution positron emission tomography (UHR-PET), especially as detector crystals become smaller and their readouts increasingly segmented. Current approaches either reject these events, reducing sensitivity, or accept them with suboptimal positioning algorithms, degrading image resolution. We present a feed forward neural network to optimize ICS event recovery by inferring the line-of-response belonging to the first Compton interaction. Our approach was validated using both Monte Carlo simulations and experimental data from the fully pixelated LabPET-II-based preclinical and brain UHR-PET scanners.Results demonstrate a 70\% to 106\% increase in sensitivity while preserving sub-millimeter spatial resolvability (down to 1.6 mm) compared to conventional methods. This ICS recovery approach is an effective solution that compensates for the lower detection efficiency of small, pixelated detectors in UHR-PET, enabling reduced scan times and lower radiation doses while largely preserving image quality.
\end{abstract}

\vspace{2pc}
\noindent{\it Keywords}: Positron Emission Tomography, Inter-Crystal Scattering, Spatial Resolution, Machine Learning, Algorithm, Neural Network

\section{Introduction}

Positron Emission Tomography (PET) continues to evolve with the development of ultra-high-resolution (UHR) systems employing highly pixelated detectors. As the crystal size is reduced to achieve higher spatial resolution, the likelihood of inter-crystal scatter (ICS) events increases significantly \citep{Shao1996}. This is because there is a low probability that the scattered photons will be reabsorbed by the same crystal after an initial Compton interaction of the 511 keV photons. The scattered photons will either escape from the detector array or be absorbed by neighboring crystals, thereby creating ICS events. If these ICS events cannot be recovered, the overall detection efficiency of single crystals is then determined by their effective photoelectric fraction, which approaches the theoretical photoelectric probability of the crystal material. This is creating a fundamental challenge for enhancing the sensitivity of modern PET scanners equipped with highly pixelated detectors. In order to recover ICS events, the identification of the first interaction point becomes crucial for accurate line-of-response (LOR) positioning.

Traditional approaches to handling ICS events face a critical dilemma: either rejecting these events, which reduces system sensitivity, or accepting them with uncertain positioning algorithms which can degrade image quality \citep{Gu2020, Kang2023}. This inaccuracy stems from the fact that identifying the first interaction point is a mathematically ill-posed inverse problem when temporal and energy resolutions are limited. Statistically, the energy deposition probability distributions for the first and subsequent interactions overlap significantly. Furthermore, because the time-of-flight difference between adjacent crystal interactions is well below the coincidence timing resolution of small-pixel detectors, temporal information cannot resolve the interaction sequence. Consequently, discerning the true sequence from noisy energy and timing measurements becomes analytically ambiguous \citep{Rafecas2003}.

While previous work, such as \citep{Lage2015}, has demonstrated the potential for sensitivity improvements through triple coincidence recovery, the challenge remains in developing accurate positioning algorithms suitable for ultra-high resolution scanners. As demonstrated by \citep{Zou2025}, the incorporation of ICS events is only viable if these positioning algorithms can preserve image spatial resolution.

Current methods for processing ICS events range from simple heuristics like the Winner-Take-All (WTA) approaches \citep{comanor1996, Shao1996, Clerk2012}, statistcal methods such as Bayesian estimation \citep{pratx2009} or maximum likelihood positioning algorithm \citep{Lage2015, gross2016} and neural network-based methods \citep{Michaud2015}. However, \citep{comanor1996} highlighted early on that WTA frequently misidentifies the first interaction in multi-scatter scenarios. More recently, \citet{surti2018} showed that while advanced positioning algorithms can improve performance, analytic methods still struggle to model the complex scattering interactions in small-pitch detectors where energy resolution is limited. As comprehensively reviewed by \citep{lee2024}, strategies for mitigating ICS have evolved from classical energy-based and Compton kinematics models to advanced statistical and artificial intelligence (AI) techniques. Recent studies \citep{sipala2024} further validated that deep learning approaches significantly outperform conventional positioning logic in such complex detector environments. Neural networks can be used for LOR estimation specifically because they can learn the non-linear mapping between the distributed energy deposition pattern and the first interaction point.

In this work, we present a complete retrofit to current state-of-the-art PET technology of the neural network method originally developed by \citep{Michaud2015}. We validate our method using both Monte Carlo simulations and experimental data from the preclinical LabPET II Mouse \citep{Gaudin2021}  and UHR \citep{Gaudin2018} PET scanner.
\section{Materials and Methods}

In this work, coincidence events are categorized into three types. Doublet events correspond to standard coincidences, in which each annihilation photon interacts with a single crystal, defining a unique LOR. Triplet events arise from ICS, where one photon undergoes Compton scattering in one crystal before being absorbed in a neighboring crystal, involving a total of three crystals. Although these events are typically discarded in scanners with fully pixelated detectors due to positioning ambiguity, they can be processed by an algorithm to identify the most probable LOR. We refer to these validated events as recovered triplets. Higher-order coincidences involving four or more crystals ($\ge \text{quadruplets}$) occur with significantly lower probability and are systematically discarded due to the unfavorable trade-off between sensitivity gain and algorithmic complexity. Finally, Combined events denote the dataset resulting from the merger of doublets and these recovered triplets, representing the optimal utilization of available data.

\subsection{Inter-Crystal Scatter Recovery Algorithm}

To efficiently discriminate the most likely Line-of-Response (LOR) amidst the stochastic noise of Compton scattering, the proposed algorithm employs a two-step strategy. This approach prioritizes the reduction of input space complexity through explicit normalization before invoking the predictive model:

\begin{enumerate}
    \item[(i)] \textbf{Geometric pretreatment} to normalize and simplify the input data, rendering the problem independent of the specific crystal pair location in the scanner.
    \item[(ii)] \textbf{A feed-forward neural network} to determine the most likely Line-of-Response (LOR) based on the energy distribution and relative geometry of the interactions.
\end{enumerate}

\subsubsection{Geometric Pretreatment}

The geometric pretreatment is a fundamental component of the proposed methodology, crucial for the performance of the neural network. By transforming raw detector coordinates into a standardized, scanner-independent canonical space, this step effectively simplifies the input feature space. It removes the spatial complexity associated with the specific location of the interaction within the scanner ring, allowing the network to focus exclusively on the energy distribution and relative geometric characteristics of Compton scattering kinematics. This normalization enables the model to more readily discriminate the correct LOR without needing to memorize the global detector architecture. This is possible due to the cylindrical symmetry of the scanner. 

The process thus follows a seven-step sequence adapted to the cylindrical scanner geometry:
\begin{enumerate}
    \item \textbf{Position quantization}: Detection localization is quantized with respect to the detector center, with continuous or discrete depth-of-interaction incorporated when available.
    \item \textbf{Energy and geometry sorting}: Detected photons are sorted by decreasing energy, and a geometric collimation is applied to address artificial backscatter artifacts arising from low energy resolution.
    \item \textbf{Transaxial symmetry removal}: Rotational and translational transformations reallign the triplet within the scanner independant framework.
    \item \textbf{Axial symmetry removal}: Similar transformations are applied along the axial dimension.
    \item \textbf{Triplet planar alignment}: The triplet is rotated to align with a standard reference plane.
    \item \textbf{Triplet scaling}: Homothety is applied to normalize the triplet axis length while preserving angles.
    \item \textbf{Normalization}: Input data is normalized to optimize neural network processing.
\end{enumerate}

\subsubsection{Neural Network Architecture}

The neural network has been optimized for the LabPET II based detector platform, consisting of:

\begin{itemize}
    \item An input layer with 6 neurons (the $x$, $y$ coordinates and energy of the two scattered singles).
    \item Three layers with 64 neurons for the clinical scanner and 48 neurons for the preclinical scanner.
    \item A single-neuron output layer that return a probability between 0 and 1 to discriminate between two candidate LORs.
\end{itemize}
The number of neurons and layers have been determined through an hyper parameter farm \citep{cometml2026}.
Hyperbolic tangent (tanh) activation functions were used for all hidden nodes. The model was trained on labeled datasets generated via Monte Carlo simulations, serving as the ground truth for event positioning. The Mouse models is train on 2.2 millions ICS cases and the UHR required a bigger dataset due to its complexity reaching a size of 3.8 millions ICS cases.

\subsection{Experimental Setup}

\subsubsection{Hardware Configuration}

Experiments were conducted using the LabPET II Mouse and UHR Brain PET scanners, which feature LYSO scintillation crystals of 1.125 × 1.125 × 10.6 mm³ and 1.125 × 1.125 × 12 mm³, respectively, arranged  in 4 x 8 arrays optically coupled to 4 x 8 monolithic arrays of avalanche photodiodes, ensuring individual pixel readouts.  The scanners have a ring architecture with a diameter and axial lenght of 78.8 mm by 50.5 mm, and 39.8 cm by 23.5 cm, respectively. These scanners achieves a spatial resolution of 0.75 mm and 1.25 mm at the center of the field of view (FOV)\citep{Gaudin2021, Gaudin2018}.

\subsubsection{Data Acquisition}\mbox{} \noindent Data were acquired in two forms:
\begin{itemize}
    \item \textbf{Monte Carlo simulations:} Using the GATE platform \citep{GATE}, accurate models of the LabPET II Mouse and UHR Brain PET scanners were used to generate simulation datasets containing $2.2 \times 10^6$ and $3.8 \times 10^6$ coincidence events, respectively.
    \item \textbf{Experimental measurements}: Point source, phantom and animal studies with the LabPET II Mouse and UYHR Brain PET scanners.
\end{itemize}

For both simulated and experimental data, we employed an energy window of 350-650 keV for photoelectric events in doublet coincidences. For triplet events, we used a singles lower energy threshold of 0 keV, comparable to the method of \citep{Abbaszadeh2023}, to capture scattered photons that deposit partial energy in multiple crystals. However, the sum of energy deposits must fall whitin the standard 350-650 keV window for the event to be recorded.

\subsubsection{In Vivo Animal Studies}
To evaluate the preservation of biological structural integrity and spatial details when incorporating recovered ICS events, an \textit{in vivo} acquisition was conducted on a $20.2~\text{g}$ mouse injected with $29~\text{MBq}$ of $^{18}\text{F-NaF}$. This tracer was selected for its high uptake in the skeletal structure, providing a high-contrast baseline for spatial resolution and contrast evaluation. The total scan duration was 75~min, starting 30~s prior to radiotracer injection to enable dynamic pharmacokinetic analyses. The acquisition was performed with an open energy window and recorded exclusively in single-event mode. Coincidences and candidate triplets were subsequently generated offline using a dedicated processing tool. A supplementary baseline acquisition using standard hardware settings and coincidence parameters was not conducted.

\subsection{Metrics}

The performance of our method was evaluated using the following metrics:

\begin{itemize}
    \item \textbf{LOR recovery rate}: Calculated as the ratio of correctly identified LORs to the total number of ICS triple coincidences for uniform sources positioned centrally in the scanners.
    \item \textbf{Sensitivity increase}: Measured as the percentage increase in detected events when including ICS triple coincidences with respect to doublets alone.
    \item \textbf{Spatial resolution}: Assessed using the \textbf{Mini Derenzo phantom} with rod diameters ranging from 1.2 to 4.8 mm.
    \item \textbf{Contrast-to-noise ratio (CNR)}: Evaluated using the \textbf{NEMA NU4 Image Quality phantom} \citep{NEMA} to assess contrast recovery and overall image quality.
\end{itemize}

All images were reconstructed using a 3D Ordered Subset Expectation Maximization (OSEM) algorithm with  analytic modeling of the coincidence detection probability in the tubes of response. All standard corrections were applied except for \textit{in vivo} studies where correction for attenuation and scatter weren't made due to the lack of a CT-scan.

\section{Results}

\subsection{Line-of-Response Recovery Performance}
The classification accuracy of the neural network was evaluated on the simulated dataset for both scanner geometries. As shown in Table \ref{tab:nn_performance}, the model achieved a Line-of-Response (LOR) recovery rate of approximately 69\% for both preclinical and clinical configurations.

\begin{table}[H]
\centering
\caption{Neural network LOR recovery performance by scanner type for a uniform cylindrical flood source ($\text{radius} = 25\text{ mm}$, $\text{height} = 50\text{ mm}$, simulated $^{18}\text{F-FDG}$ activity of $100\text{ kBq}$) centered in the field of view.}
\label{tab:nn_performance}
\begin{tabular}{lc}
\hline
\textbf{Scanner Model} & \textbf{LOR Recovery Rate (\%)} \\
\hline
LabPET II Mouse & 69.0 \\
UHR Brain & 68.7 \\
\hline
\end{tabular}
\end{table}

\subsection{Sensitivity Analysis}

\begin{figure}[H]
    \centering
    \includegraphics[width=0.8\textwidth]{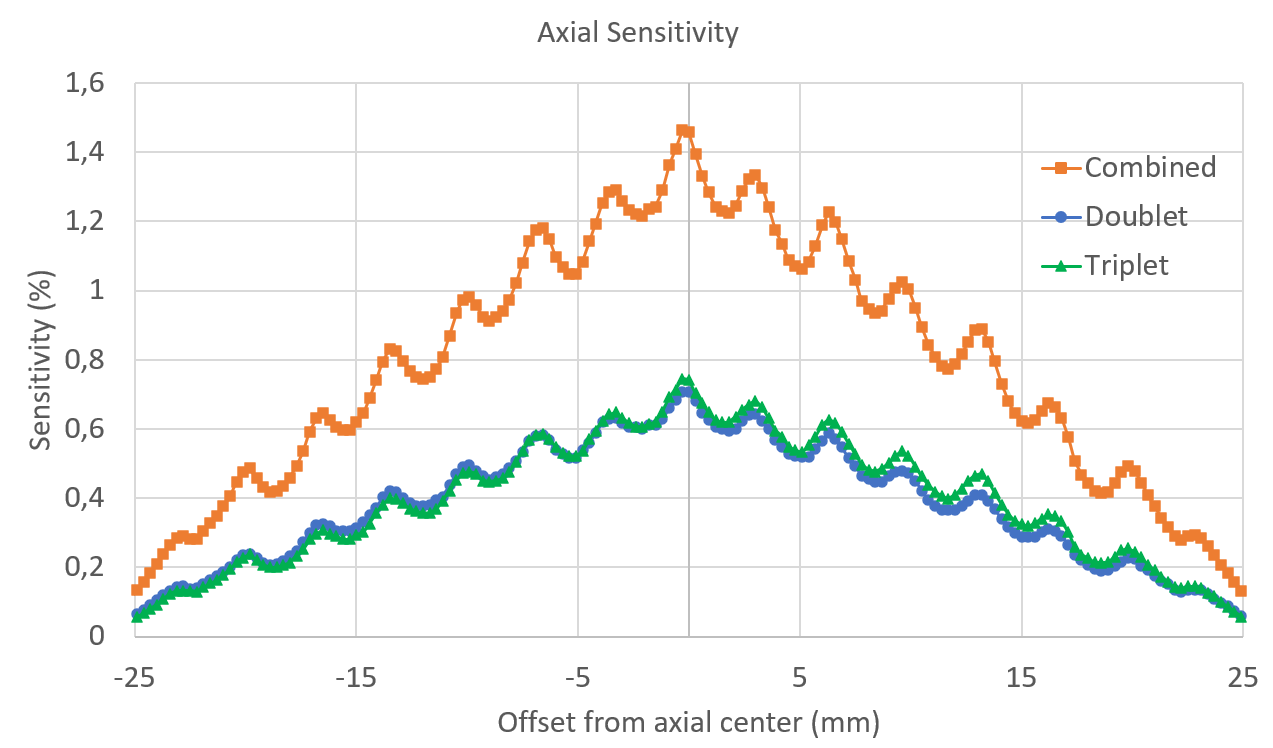}
    \caption{Axial sensitivity profiles measured on the LabPET II Mouse scanner. The blue curve represents the Doublets onlys method (individual interactions within 350--650~keV). The orange curve represents the proposed Combined method, which recovers ICS events by validating the summed energy of scattered interactions. The green curve represent only the ICS event that are candidate to recovery in the proposed method. The inclusion of these events results in a 106\% sensitivity gain at the center of the axial field of view.}
    \label{fig:Sensitivity}
\end{figure}

The axial sensitivity profiles comparing the standard Doublet dataset and the proposed Combined dataset are presented in Figure~\ref{fig:Sensitivity}. The Combined dataset, which integrates doublet events with recovered ICS events, achieved a peak absolute sensitivity of 1.46\% at the center of the field of view (CFOV) on the LabPET II Mouse scanner. In comparison, the traditional Doublet method yielded an absolute sensitivity of 0.71\% when applying the standard 350-650 keV energy window.

This corresponds to a sensitivity increase of \textbf{106\%}. This substantial gain is directly attributable to the proposed energy windowing strategy, which recovers scattered photons via summed energy validation rather than rejecting them based on individual interaction thresholds.

\subsection{Image Quality Assessment}

\subsubsection{Contrast Recovery}
To evaluate the impact of ICS recovery on contrast recovery and quantitative accuracy, scans of the NEMA NU4 Image Quality phantom were reconstructed using the OSEM algorithm.

\begin{figure}[H]
    \centering
    \begin{subfigure}[t]{0.32\textwidth}
        \centering
        \includegraphics[width=\linewidth]{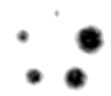}
        \caption{Doublet}
        \label{fig:rods_doublet}
    \end{subfigure}
    \hfill
    \begin{subfigure}[t]{0.32\textwidth}
        \centering
        \includegraphics[width=\linewidth]{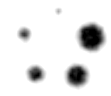}
        \caption{Combined}
        \label{fig:rods_combined}
    \end{subfigure}
    \hfill
    \begin{subfigure}[t]{0.32\textwidth}
        \centering
        \includegraphics[width=\linewidth]{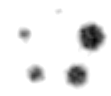}
        \caption{Triplet}
        \label{fig:rods_triplet}
    \end{subfigure}
    
    \vspace{0.5cm} 

    \begin{subfigure}[b]{0.8\textwidth}
        \centering
        \includegraphics[width=\linewidth]{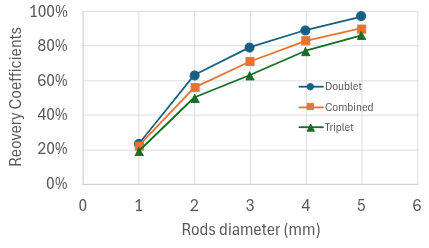}
        \caption{Quantitative Recovery Coefficients}
        \label{fig:graph_rc}
    \end{subfigure}
    
    \caption{Image quality evaluation using the NEMA NU4 phantom (UHR scanner, OSEM 16 subsets, 6 iterations). \textbf{(a)-(c)} Transaxial views of the rod sector for Doublet, Triplet, and Combined datasets respectively. Note the preservation of rod separation in the Combined dataset compared to Doublet. \textbf{(d)} Recovery coefficients as a function of rod diameter. The Combined dataset (orange) maintains quantitative accuracy within 10\% of the standard Doublet dataset (blue) across all rod sizes.}
    \label{fig:NEMA_Results}
\end{figure}

Visual inspection of the rod sectors (Figure~\ref{fig:NEMA_Results}a-c) confirms that the proposed Combined method preserves the structural definition of the rods compared to the Doublet only reconstruction. Figure~\ref{fig:NEMA_Results}d quantifies the recovery coefficients (RC) as a function of rod diameter (1--5 mm). 

As expected, the Doublet only method yields the highest recovery coefficients due to the accurate positioning of photoelectric interactions. While the Triplet method (green curve) shows a reduction in RC due to the inherent spatial uncertainty of recovered Compton events, the Combined method (orange curve) maintains recovery coefficients very close to the reference benchmark  involving Doublet only coincidences. The measured loss in RC is consistently less than 10\% across all rod sizes compared to the Doublet method. This demonstrates that mixing recovered ICS events with standard events achieves a 106\% sensitivity gain with only minor degradation of the contrast or quantitative accuracy of small structures.

\subsubsection{Uniformity and Spill-over Performance}

\begin{figure}[htbp]
    \centering
    
    % --- LIGNE 1 : SPILL-OVER ---
    \begin{subfigure}[b]{0.32\textwidth}
        \centering
        \includegraphics[width=\linewidth]{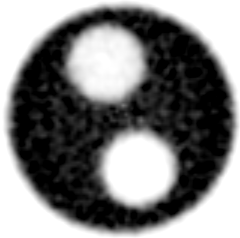}
        \caption{Doublet (Spill-over)}
    \end{subfigure}
    \hfill
    \begin{subfigure}[b]{0.32\textwidth}
        \centering
        \includegraphics[width=\linewidth]{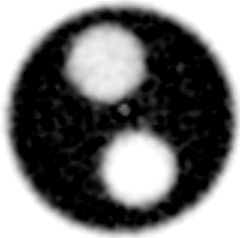}
        \caption{Combined (Spill-over)}
    \end{subfigure}
    \hfill
    \begin{subfigure}[b]{0.32\textwidth}
        \centering
        \includegraphics[width=\linewidth]{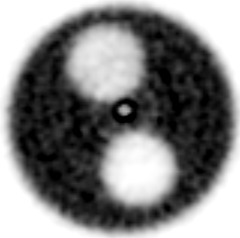}
        \caption{Triplet (Spill-over)}
    \end{subfigure}
    
    \vspace{0.5cm}
    
    % --- LIGNE 2 : UNIFORMITY ---
    \begin{subfigure}[b]{0.32\textwidth}
        \centering
        \includegraphics[width=\linewidth]{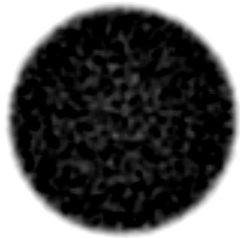}
        \caption{Doublet (Uniformity)}
    \end{subfigure}
    \hfill
    \begin{subfigure}[b]{0.32\textwidth}
        \centering
        \includegraphics[width=\linewidth]{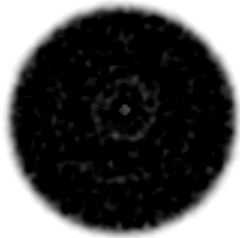}
        \caption{Combined (Uniformity)}
    \end{subfigure}
    \hfill
    \begin{subfigure}[b]{0.32\textwidth}
        \centering
        \includegraphics[width=\linewidth]{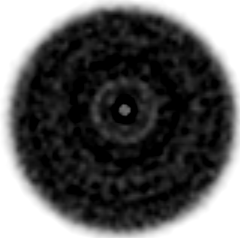}
        \caption{Triplet (Uniformity)}
    \end{subfigure}
    
    \caption{NEMA NU4 image quality assessment (UHR scanner). \textbf{Top Row:} Spill-over phantom sections containing water and air chambers. \textbf{Bottom Row:} Uniformity phantom sections. Visual inspection confirms that the Combined method maintains background homogeneity comparable to the standard Doublet method.}
    \label{fig:nema_unif_spill}
\end{figure}

\begin{table}[H]
    \centering
    \caption{Image quality metrics using NEMA NU4 phantom (UHR scanner, OSEM 16 subsets, 6 iterations).}
    \label{tab:nema_uniformity_spilloverr}
    \begin{tabular}{lccc}
        \toprule
        \textbf{Method} & \textbf{Uniformity} & \textbf{Water Spill-over} & \textbf{Air Spill-over} \\
         & (\%STD) & (Ratio) & (Ratio) \\
        \midrule
        Doublet  & 9.45 & $0.04 \pm 0.01$ & $0.17 \pm 0.03$ \\
        Triplet  & 10.35 & $0.10 \pm 0.02$ & $0.25 \pm 0.04$ \\
        Combined & \textbf{9.44} & $0.07 \pm 0.02$ & $0.21 \pm 0.03$ \\
        \bottomrule
    \end{tabular}
\end{table}

Table~\ref{tab:nema_uniformity_spilloverr} and Figure~\ref{fig:nema_unif_spill} present the uniformity and spill-over results. Uniformity, expressed as the percentage standard deviation (\%STD) in a uniform region, serves as a proxy for image noise texture.

Remarkably, the proposed Combined method achieved a uniformity of \textbf{9.44\%}, which is virtually identical to the standard Doublet method (9.45\%). This result indicates that the inclusion of ICS events does not degrade the signal-to-noise ratio in uniform regions. While recovered ICS events inherently carry higher spatial uncertainty (as evidenced by the Triplet-only uniformity of 10.35\%), the 73\% increase in count statistics provided by the Combined method effectively counterbalances this uncertainty, resulting in a smooth, high-quality background.

Regarding spill-over accuracy, the Combined method yielded a water spill-over ratio of $0.07 \pm 0.02$, representing a minor increase compared to the Doublet baseline ($0.04 \pm 0.01$). This slight elevation in the cold region background is expected due to the "smoothing" effect of the fraction of mispositioned ICS events. However, it remains significantly lower than the Triplet-only result ($0.10 \pm 0.02$) and is well within acceptable limits for preclinical quantification, particularly given the benefits in sensitivity.

\subsubsection{Spatial Resolution and Efficiency Analysis}

To validate the practical impact of the sensitivity gain, we performed an iso-count(approximately the same number of event in both acquisition)comparison using the Mini Derenzo phantom and the UHR Brain PET scanner. The goal was to determine if the proposed method allows for a reduction in scan time while maintaining spatial resolution.

\begin{figure}[H]
    \centering

    \begin{subfigure}[t]{0.32\textwidth}
        \centering
        
        \includegraphics[width=\linewidth]{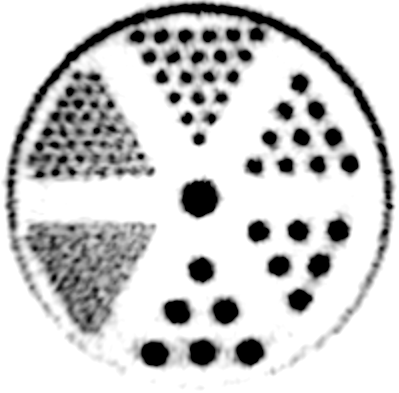}
        \caption{Doublet (28 min scan)}
        \label{fig:derenzo_dbl}
    \end{subfigure}
    \hspace{1cm}
    \begin{subfigure}[t]{0.32\textwidth}
        \centering
        \includegraphics[width=\linewidth]{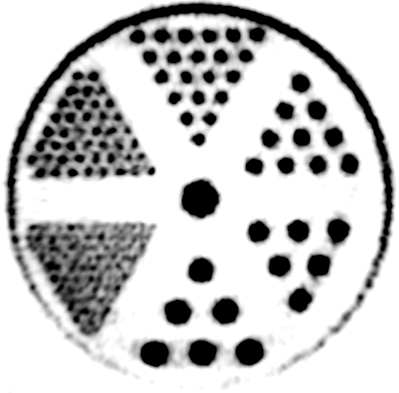}
        \caption{Combined (14 min scan)}
        \label{fig:derenzo_comb}
    \end{subfigure}
    
    \vspace{0.5cm} 
    
    \begin{subfigure}[b]{0.8\textwidth}
        \centering
        \includegraphics[width=\linewidth]{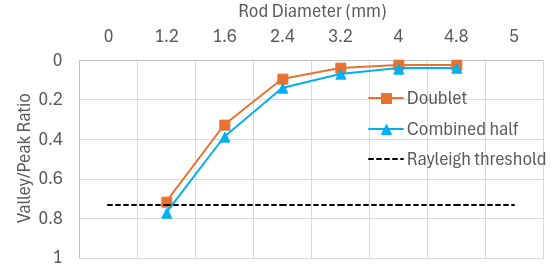}
        \caption{Quantitative resolvability analysis}
        \label{fig:rayleigh_graph}
    \end{subfigure}
    
    \caption{Iso-count spatial resolution assessment using the Mini Derenzo phantom. \textbf{(a)} Reference image reconstructed using Doublet only events from a 28-minute acquisition ($\sim$133M counts). \textbf{(b)} Image reconstructed using the Combined dataset from a 14-minute acquisition ($\sim$134M counts). Despite the 50\% reduction in scan time, the Combined dataset achieves comparable visual resolution. \textbf{(c)} Valley-to-peak ratio analysis. The combined dataset (blue curve) satisfies the Rayleigh criterion ($<0.7$) down to 1.6 mm, matching the performance of the Doublet dataset (orange curve)}
    \label{fig:Derenzo_Res}
\end{figure}

As illustrated in Figure~\ref{fig:Derenzo_Res}, the standard Doublet image was reconstructed from a full 28-minute acquisition (133 million counts), whereas the Combined image was reconstructed from only 14 minutes of data (134 million counts). By leveraging the 73\% sensitivity increase, the Combined method achieved equivalent statistics in half the time.

Visual inspection (Figure 4a--b) reveals that the 14-minute Combined reconstruction produces a resolvability pattern virtually identical to the 28-minute standard reference. Rods are clearly resolved down to the 1.6~mm sector. 

To objectively assess rod separation and eliminate subjective visual bias, we applied the Rayleigh criterion to the line profiles drawn across each rod sector of the Mini Derenzo phantom. Historically defined in optics, this criterion considers two adjacent intensity peaks resolved if the valley-to-peak (V/P) intensity ratio between them falls below a standard threshold of approximately 0.735 (corresponding to an intensity dip of at least 26.5\%). This quantitative metric provides a robust, standardized benchmark to strictly define the spatial resolution limit across different acquisition durations and datasets.

As presented in Figure 4c, this quantitative analysis confirms our visual observations. The performance curves is above the criterion for all rod sizes $\ge 1.6~\text{mm}$. At the 1.2~mm limit, which approaches the physical resolution of the scanner, a minor reduction in peak separation is observed for the Combined method. However, this slight trade-off at the extreme resolution limit is outweighed by the ability to halve the examination duration (or the activity) without compromising the visibility of clinically relevant structures.

\subsection{In Vivo Validation}

\begin{figure}[H]
    \centering
    \begin{subfigure}[b]{0.35\textwidth}
        \centering
        \includegraphics[width=\linewidth]{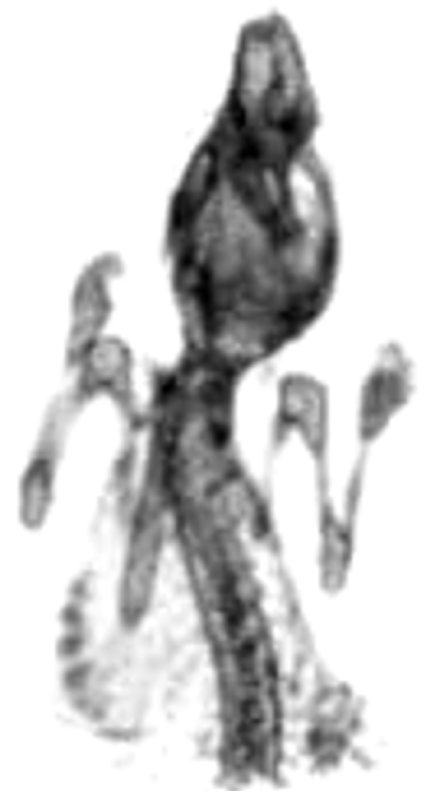}
        \caption{Doublet}
        \label{fig:souris_doublet}
    \end{subfigure}
    \hspace{2cm}
    \begin{subfigure}[b]{0.35\textwidth}
        \centering
        \includegraphics[width=\linewidth]{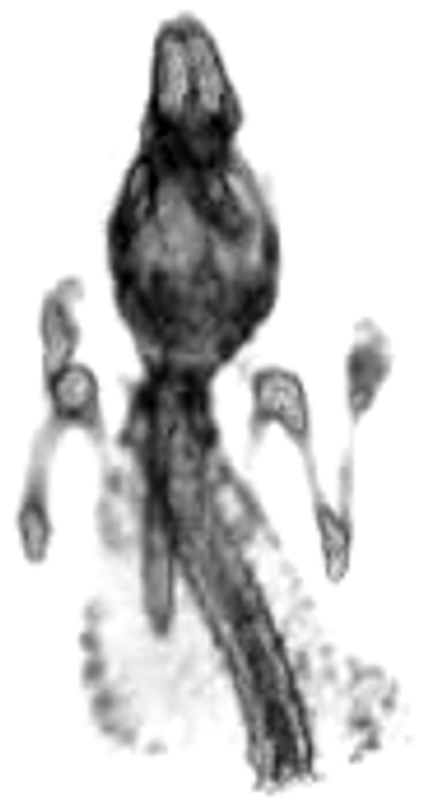}
        \caption{Combined}
        \label{fig:souris_combined}
    \end{subfigure}
    
    \caption{Comparative Na$^{18}$F PET images of a 20.2 g mouse (29 MBq injected activity). (a) Reconstruction using Doublet only events. (b) Reconstruction using the proposed Combined method. Although a minor trade-off in edge sharpness is visible in the Combined reconstruction, the method maintains structural consistency with the dataset without ICS while utilizing a significantly larger dataset.}
    \label{fig:comparaison_souris}
\end{figure}

Figure \ref{fig:comparaison_souris} presents a qualitative comparison between the Doublet only reconstruction and the proposed Combined dataset. Visual inspection confirms that the Combined dataset preserves the overall anatomical integrity of the subject. While a subtle smoothing of high-frequency spatial details can be observed, particularly in the separation of vertebral segments, the primary skeletal structures remain clearly identifiable. This indicates that the inclusion of ICS events achieves a substantial sensitivity gain while maintaining a spatial resolution sufficient for the localization of anatomical features.

\subsection{Improvement in Temporal Resolution for Dynamic Studies}
\begin{figure}[H]
    \centering
    \begin{subfigure}[b]{0.32\textwidth}
        \centering
        \includegraphics[width=\linewidth]{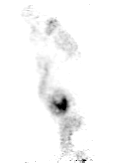}
        \caption{Doublet (8 min acq.)}
    \end{subfigure}
    \hspace{2cm}
    \begin{subfigure}[b]{0.32\textwidth}
        \centering
        \includegraphics[width=\linewidth]{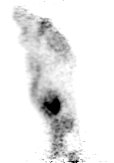}
        \caption{Combined (4 min acq.)}
    \end{subfigure}
    
    \vspace{0.5cm}
    
    \begin{subfigure}[b]{0.32\textwidth}
        \centering
        \includegraphics[width=\linewidth]{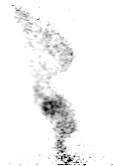}
        \caption{Doublet (2 min acq.)}
    \end{subfigure}
    \hspace{2cm}
    \begin{subfigure}[b]{0.32\textwidth}
        \centering
        \includegraphics[width=\linewidth]{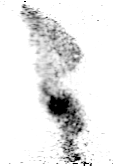}
        \caption{Combined (1 min acq.)}
    \end{subfigure}
    
    \caption{Impact of ICS recovery on temporal resolution for dynamic imaging ($^{18}$F-FDG, 20.2 g mouse). Top row: Comparison of a standard duration frame (a) versus a half-duration frame using the Combined method (b) at 10 min post-injection. Bottom row: Comparison during the early uptake phase (2 min post-injection), showing standard (c) versus half-duration Combined reconstruction (d). The Combined dataset allows for a 50\% reduction in frame duration while maintaining image quality comparable to the standard Doublet dataset.}
    \label{fig:dynamic_study}
\end{figure}

Beyond static image quality, the sensitivity gain provided by the Combined dataset has direct implications for dynamic PET studies, where temporal resolution is critical for accurate pharmacokinetic modeling. Figure \ref{fig:dynamic_study} illustrates the potential for reducing frame duration without compromising image quality.

We compared Doublet only reconstructions with Combined reconstructions acquired over half the duration ($T_{combined} = 0.5 \times T_{doublet}$). In both high-statistics scenarios (10 min post-injection) and low-statistics early uptake phases (2 min post-injection), the Combined method at half-duration yielded images visually equivalent to the full-duration standard benchmarks.

Specifically, the Combined reconstruction of a 1-minute frame (Fig. \ref{fig:dynamic_study}d) exhibits noise levels and structural definition comparable to the 2-minute standard acquisition (Fig. \ref{fig:dynamic_study}c). This suggests that the inclusion of ICS events allows for a twofold increase in temporal sampling rates. This capability is particularly advantageous for capturing rapid radiotracer kinetics in the early phases of distribution, effectively doubling the temporal resolution available for generating time-activity curves.

\subsection{Computational Performance}
To evaluate the execution speed of the proposed ICS recovery algorithm, deterministic profiling was performed using the \texttt{cProfile} tool on an Intel i9-11900K workstation. The core processing pipeline, encompassing the geometric pretreatment and neural network inference, achieved an average processing throughput of approximately 6,300 events per second. The profiling analysis revealed that processing time was primarily bounded by file input/output (I/O) operations when reading raw event data from external storage, whereas the linear algebra transformations and network matrix operations accounted for a minor fraction of the total execution time.
\section{Discussion}

\subsection{Balancing Sensitivity and Spatial Resolution}
The primary objective of this work was to unlock the sensitivity reserve of Inter-Crystal Scatter (ICS) events without compromising the high-resolution capabilities of the scanners based on the LabPET II pixelated detection platform. The results demonstrate that the proposed machine learning-based positioning, combined with geometric pre-processing, successfully achieves this balance.

Specifically, nearly 70\% of the recovered ICS events are assigned to the correct crystal pair (Table~\ref{tab:nn_performance}), thereby contributing valid spatial information for the image reconstruction. While the remaining 31\% of mispositioned events introduce a bounded spatial uncertainty, this trade-off is necessary to access the significant sensitivity reserve contained within ICS data.

Most importantly, the NEMA and Derenzo phantom studies confirm that these mispositioned events have little impact for most practical purposes. The uniformity analysis showed no significant increase of noise, suggesting that the statistical gain effectively suppresses the spatial noise introduced by positioning errors. While a minor loss in contrast recovery ($<10\%$) and resolvability at the 1.2 mm limit was observed, the resolution for structures $\ge$ 1.6 mm remains preserved. This indicates that the method remains suitable for imaging major anatomical structures in small animals, provided that fine sub-millimeter details below this limit are not the primary target of the analysis.

\subsection{Implications for Low-Dose and Dynamic Imaging}
The most significant finding of this study is the translation of sensitivity gains into temporal efficiency. As demonstrated in the iso-count Derenzo experiment (Figure~\ref{fig:Derenzo_Res}) and the \textit{in vivo} dynamic study (Figure~\ref{fig:dynamic_study}), the Combined method allows for a twofold reduction in scan time in the preclinical setting.

This has two major implications for research. First, it enables high-temporal-resolution dynamic imaging. By maintaining image quality with shorter frames, the method allows for more precise pharmacokinetic modeling of rapid tracer uptake. Second, it offers a pathway for dose reduction. Achieving comparable image quality with twice as much counts implies that the injected activity could essentially be halved, reducing radiation exposure to study subjects and staff.

\subsection{Comparison with Previous Work}
The proposed machine learning approach represents a distinct evolution from earlier algorithmic methods for ICS recovery. Classical physics-based approaches, such as the one proposed by \citep{Rafecas2003}, relied on Compton kinematics and energy discrimination, achieving interaction sequence identification fractions of approximately 60\%. However, their model assumed a dual-layer detector architecture with an optimal energy resolution of 15\%, a configuration that inherently provides stronger physical constraints for event decoding. In contrast, even under the significantly degraded energy resolution conditions typical of highly pixelated single-layer detectors, our neural network leverages geometric learning to achieve a Line-of-Response recovery accuracy of nearly 70\%. The key advantage here is the network's ability to learn complex spatial dependencies that are difficult to model analytically, particularly in small-pitch detectors where energy resolution is a severe limiting factor.

Furthermore, this work significantly improves upon the initial neural network implementation by \citep{Michaud2015}. While their method achieved a sensitivity increase of 54\%, our optimized architecture yielded a 73\% gain on the clinical UHR geometry and over 100\% on the preclinical scanner. This performance leap validates the effectiveness of the refined geometric pre-treatment made during the reassignment of the cristal position into the augmented space and the adaptation of the network architecture to specific scanner geometries.

\subsection{Computational Efficiency and Real-Time Implementation Feasibility}
Although the current processing throughput of 6,300 events per second is insufficient for real-time acquisition streams, this baseline reflects an unoptimized, single-threaded Python research prototype designed for post-acquisition offline processing. Consequently, this performance rate does not represent the intrinsic speed limits of the algorithm nor its potential in a clinical setup.

The low computational overhead of the core algorithm strongly supports the feasibility of real-time processing. The geometric pretreatment relies exclusively on basic linear algebra operations (rotation and scaling), and the neural network architecture comprises approximately 1.5 million parameters requiring deterministic matrix multiplications. These operations are highly parallelizable and well-suited for hardware acceleration. As demonstrated by previous Field-Programmable Gate Array (FPGA) implementations for PET event processing \citep{geoffroy2015}, embedding this logic directly onto a Field-Programmable Gate Array at the detector front-end would bypass external memory I/O bottlenecks and enable online, real-time ICS recovery.

\subsection{Limitations and Future Perspectives}
While the proposed method maintains spatial resolution for structures $\ge$ 1.6 mm, a slight performance reduction was observed at the 1.2 mm resolution limit (Figure~\ref{fig:Derenzo_Res}). This suggests that for specific applications requiring maximum spatial resolution at the detector's physical boundary, the standard Doublet method may still be preferable, albeit at the cost of longer scan times.

Future work will focus on two main axes to address this limitation and further enhance performance. First, the current model focuses on single Compton scattering events. However, in high-resolution pixelated detectors with small crystal pitches, multiple sequential Compton interactions become increasingly probable. Extending the network architecture to identify higher order interaction sequences involving three or more crystals could unlock an additional sensitivity reserve.

Second, we aim to integrate physics-based constraints directly into the learning process. The current feed-forward architecture relies primarily on data-driven patterns. The development of Physics-Informed Neural Networks (PINNs) could allow for the incorporation of Compton kinematics and energy conservation laws as components of the loss function. This hybrid approach would penalize physically impossible predictions, potentially improving the positioning accuracy for ambiguous events and further reducing the contrast trade-off observed in this study.
\section{Conclusion}

In this work, we have demonstrated that recovering Inter-Crystal Scatter (ICS) events using a geometry-aware neural network significantly enhances the performance of ultra-high resolution PET systems. By turning discarded inter-crystal scatter events into valid coincidences, our method achieves a 73\% sensitivity gain on the UHR scanner and 106\% on the LabPET mouse scanner with minimal impact on spatial resolution (preserving resolvability for structures $\ge 1.6$~mm) or contrast (<10\% reduction).

The most critical outcome of this study is the translation of this sensitivity reserve into practical imaging benefits. We validated that the proposed method enables up to a twofold reduction in scan time or equivalently, injected dose without compromising diagnostic image quality. Furthermore, it unlocks new capabilities for dynamic imaging by supporting higher temporal sampling rates essential for pharmacokinetic modeling. Consequently, this approach offers a versatile solution to the sensitivity challenges of pixelated detectors, paving the way for more efficient and quantitative preclinical imaging.

\section*{Acknowledgments}

This work was supported in part by the Discovery Grants RGPIN-2025-06139 of the Natural Sciences and Engineering Research Council of Canada (NSERC), and the Acuity-QC Consortium, Project FACS-008, funded by the Quebec Ministry of Economy and Innovation via CQDM. The Digital Research Alliance of Canada provided the computing resources under grants 2630, 4063, 5036. The authors would like to thank Jean-François Beaudoin, Vincent Doyon, Maxime Toussaint, Maxime Gaudreault, Christian Thibaudeau, Jacob Lambert, Sophie Carneiro Esteves and Étienne Auger for their assistance with animal handling and data acquisition, as well as the Digital Research Alliance of Canada for providing the computational resources necessary for neural network training.

\bibliography{references}

\end{document}